\documentclass[letterpaper, 10 pt, conference]{ieeeconf}
\IEEEoverridecommandlockouts
\usepackage{gensymb}
\usepackage{graphicx}
\usepackage{amsmath}
\usepackage{cite}
\usepackage{xcolor}
\usepackage{tabularray}
\usepackage{booktabs}
\usepackage{caption}
\usepackage[T1]{fontenc}
\usepackage[utf8]{inputenc}
\usepackage{makecell}
\usepackage{multirow}

\title{\LARGE \bf
Optimizing Locomotor Task Sets in Biological Joint Moment Estimation for Hip Exoskeleton Applications}

\author{Jimin An$^{1}$, Changseob Song$^{1}$, Eni Halilaj$^{1}$ and Inseung Kang$^{1}$
\thanks{$^{1}$J. An, C. Song, E. Halilaj, and I. Kang are with the Department of Mechanical Engineering, Carnegie Mellon University, Pittsburgh, PA 15213, USA. Corresponding author: Jimin An ({\tt\footnotesize jimina@andrew.cmu.edu}).}
}

\begin{document}

\maketitle
\thispagestyle{empty}
\pagestyle{empty}

\begin{abstract}
Accurate estimation of a user's biological joint moment from wearable sensor data is vital for improving exoskeleton control during real-world locomotor tasks. However, most state-of-the-art methods rely on deep learning techniques that necessitate extensive in-lab data collection, posing challenges in acquiring sufficient data to develop robust models. To address this challenge, we introduce a locomotor task set optimization strategy designed to identify a minimal, yet representative, set of tasks that preserves model performance while significantly reducing the data collection burden. In this optimization, we performed a cluster analysis on the dimensionally reduced biomechanical features of various cyclic and non-cyclic tasks. We identified the minimal viable clusters (i.e., tasks) to train a neural network for estimating hip joint moments and evaluated its performance. Our cross-validation analysis across subjects showed that the optimized task set-based model achieved a root mean squared error of 0.30±0.05 Nm/kg. This performance was significantly better than using only cyclic tasks (p<0.05) and was comparable to using the full set of tasks. Our results demonstrate the ability to maintain model accuracy while significantly reducing the cost associated with data collection and model training. This highlights the potential for future exoskeleton designers to leverage this strategy to minimize the data requirements for deep learning-based models in wearable robot control.
\end{abstract}

\begin{keywords}
Exoskeleton, Deep Learning, Biological Joint Moment Estimation, Dimensionality Reduction, Cluster Analysis
\end{keywords}

\section{Introduction}
Lower-limb exoskeletons have been shown to effectively enhance human mobility, thereby improving quality of life \cite{YoungExpansion}. These robotic systems can augment users and enhance gait performance, for example, by reducing energy expenditure in healthy individuals \cite{Slade2022Nature} and improving spatiotemporal gait metrics in patient populations \cite{hemiparetic}. Among different lower-limb exoskeletons, hip exoskeletons present a particularly compelling value proposition due to the critical role of the hip in generating substantial mechanical power during locomotion. Unlike the ankle joint, which benefits from an efficient muscle-tendon mechanism, the hip is mechanically less efficient, requiring greater muscle work (i.e., energy) to complete locomotor tasks \cite{Sawicki2009-xd}.

Control strategies for robotic exoskeletons have significantly advanced over the past decade. Initially relying on heuristic, rule-based controllers utilizing onboard wearable sensors \cite{CollinsWholeLeg, YoungTiming}, recent approaches now directly leverage the user's physiological state, such as biological joint moments, to control the exoskeleton more effectively \cite{Gasparrif_TNSRE_2019}. Biological torque controllers directly estimate the user's biological joint moments and apply a reshaped, scaled envelope as the assistance profile \cite{Molinaro_BioRob_2020, Molinaro_ICRA_2023}. This method leverages deep learning techniques with wearable sensing as model input and has demonstrated significant performance benefits \cite{Molinaro_Nature, Molinaro_Sci_Robot}. The range of activities that can be assisted by exoskeletons has expanded significantly, progressing from simple continuous walking \cite{10086574, Carmago_JBiomech_2021} to dynamic transitions between locomotion modes \cite{Molinaro_TMRB_2022, Molinaro_Sci_Robot}, and now including various non-cyclic movements \cite{Improving, Molinaro_Nature}. These advancements hold the promise of enabling more natural hip exoskeleton assistance during daily locomotor tasks.

However, a significant bottleneck of deep learning approaches is their data-intensive nature, requiring large datasets that comprehensively represent various activity modes, along with ground-truth joint moments obtained through motion capture. Studies have demonstrated that user-independent models typically require data from at least a dozen participants completing multiple tasks \cite{Kang_TBME_2022, 10086574}, with some relying on even larger groups and more diverse tasks \cite{Molinaro_Sci_Robot, Molinaro_Nature}. While comprehensive datasets are essential for ensuring reliable model performance across various activities, the associated costs and resources are substantial. Furthermore, for patient populations, such data collection can be even more challenging due to physical limitations, time constraints, and logistical barriers. These challenges highlight the necessity for an automated solution that reduces the burden of data collection while not compromising model performance.

\begin{figure*}[t!]
    \centering
    \includegraphics[width=1.0\textwidth]{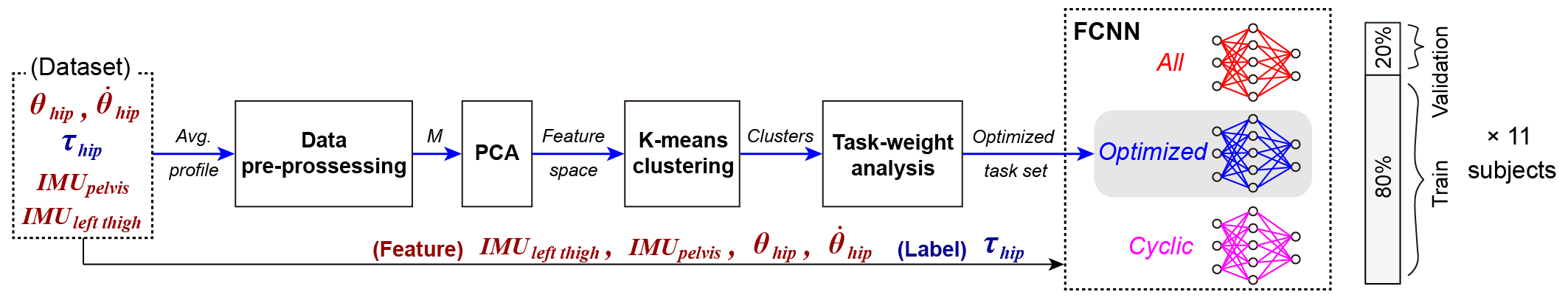}
    \caption{
        Task Selection and Model Training. Average joint moment, angle, and velocity profiles were used for principal component analysis (PCA), k-means clustering, and task-weight analysis to determine a smaller optimized set of tasks that were representative of the full set. The time-series data for each training dataset condition (all tasks, optimized tasks, and cyclic tasks) were used for the fully connected neural network (FCNN) model training with leave-one-subject-out cross-fold validation.
        }
\end{figure*}

Few studies have verified that optimal model performance can be achieved with data from a reduced set of tasks \cite{Molinaro_Nature, Improving}. These studies employed an iterative forward task selection approach, incrementally adding tasks to the training dataset until further additions no longer improved model performance. While these results suggest that a full dataset is not always necessary during training, they do not clarify which tasks are most relevant and useful or explain why certain tasks are more effective in developing user-independent joint moment estimators.
Without a deeper analysis beyond model performance, these approaches provide limited insight into the relationships between tasks and their biomechanical relevance. 
This gap emphasizes the need for a systematic approach to identify a minimal task set that ensures reliable model training and performance while reducing data collection requirements.

Given the similarities among human movement profiles across various locomotor tasks, collecting data with finely discretized task separation may not always be necessary. Instead, grouping tasks based on shared biomechanical features might be a better strategy for improving efficiency and applicability in various contexts. Prior studies have applied dimensionality reduction and clustering methods to biomechanical data analysis for applications such as locomotion phase classification \cite{locomotionphase, stanceswing}, kinematic pattern recognition \cite{kinematicpattern}, gait differentiation \cite{gaitdifferentiation}, and movement analysis \cite{moveAnalysis1, moveAnalysis2, moveAnalysis3}. Building on prior literature, grouping tasks using a clustering approach based on shared biomechanical features offers a systematic solution for identifying redundant tasks that provide similar information. By selecting representative tasks from each cluster and training the model on these tasks, we can significantly reduce data collection efforts while preserving essential biomechanical patterns needed for accurate joint moment estimation.

In this work, we propose a strategy for optimizing locomotor task selection through clustering analysis. We hypothesized that a model trained on an optimized task set would achieve hip joint moment estimation comparable to a model trained on the full dataset. Our underlying rationale is that the latent space representation of these tasks likely exhibits significant overlap for certain tasks due to their shared biomechanical features. To test this hypothesis, we trained a neural network using three different datasets to evaluate the effectiveness of our task set reduction: 1) the full set of tasks, 2) an optimized reduced set of tasks, and 3) only cyclic tasks. The model estimated hip joint moments using sensor data commonly available for hip exoskeletons, such as encoders and inertial measurement units. Our study results contribute to the field by demonstrating that a biological torque controller for hip exoskeletons can be developed using data from fewer tasks, while also identifying key tasks for future data collection efforts.

\section{Methods}
\subsection{Dataset and Data Pre-Processing}
We used an open-source biomechanics dataset of 12 healthy young individuals (7 males, 5 females, age of 21.8$\pm$3.2 years, height of 176.7$\pm$8.6 cm, and weight of 76.9$\pm$14.4 kg) performing various locomotor tasks \cite{Scherpereel_SciData_2023}. These tasks included 10 cyclic and 17 non-cyclic tasks. The dataset included kinematics (joint angles and velocities), kinetics (ground reaction forces, joint moments, and joint power), and wearable sensor data. For dimensionality reduction, we focused on unilateral hip joint angles, velocities, and moments in the sagittal plane. These features, combined with pelvis and thigh inertial measurement unit (IMU) data, were also used to train a user-independent model for hip moment estimation in exoskeletons.

We used principal component analysis (PCA) to reduce the dimension of our data. We constructed an input matrix, $M$, through a pre-processing pipeline (Fig. 1). The unilateral hip joint moment, angle, and velocity, pre-segmented and averaged by gait or movement cycle, were linearly interpolated to standardize their lengths while preserving overall patterns. These processed vectors were concatenated to represent the biomechanical features for each subject-task combination, and all such vectors were aggregated to form the matrix, $M$. During this process, seven non-cyclic tasks (meander, poses, push, obstacle walk, start stop, tug of war, and twister) were excluded due to their atypical movement profiles, which could not be segmented into uniform gait or movement cycles. The remaining tasks were categorized into eight cyclic and 12 non-cyclic tasks. We excluded data from one subject due to insufficient cyclic task trials, ensuring balanced task representation and minimizing potential biases in analyses.

\subsection{PCA, Clustering, and Task Selection}
 By reducing high-dimensional biomechanical data, PCA \cite{moveAnalysis2} enhanced the interpretability of relationships among task datasets. We derived principal components and assessed their explained variance to determine the optimal number of components. This process balanced variance representation and component counts, ensuring an adequate representation of the data's variability for the clustering process.

 We applied k-means clustering  \cite{stanceswing} to group tasks by biomechanical feature similarities in the PCA-derived feature space (Fig. 1). We first performed a silhouette score analysis to determine the number of clusters. The silhouette score measures both inter-cluster and intra-cluster distances, indicating how well-defined the clusters are. The $K$ (i.e., number of clusters) that yielded the highest silhouette score was selected as the optimal $K$.

Since each cluster contained different types of tasks, we developed a task-weight analysis to calculate, $R$, the representativeness of a task for a cluster using Eq. 1,

\begin{equation} 
R = \frac{A}{B} \times \frac{A}{C} \times \frac{S}{11} \times w
\end{equation}
where $A$ is the number of a specific task's data points in a cluster, $B$ is the total number of data points in that cluster, $C$ is the total number of data points for that task across all clusters, and $S$ is the number of unique subjects contributing task data in the cluster. $w$ is a weight assigned to each task based on its popularity in the research field and the ease of collecting task data, reflecting the data collection difficulty. Using Eq. 1, we accounted for the proportion of data, the diversity of subjects, and the difficulty associated with data collection in our task selection process. The representative tasks for each cluster identified through this analysis constituted our set of optimized tasks for model training.

\subsection{Optimized Task Set Performance Validation}
We implemented a fully connected neural network (FCNN) using PyTorch (Python v3.8) to predict a unilateral hip joint moment from biomechanical and IMU data. The network architecture was comprised of an input layer, three hidden layers, and an output layer. The input layer had a 14-dimensional feature vector corresponding to the input features (Fig. 1). Each hidden layer contained 50 nodes. We applied batch normalization after each linear transformation to enhance training stability and convergence speed. We used the rectified linear unit activation function followed by a dropout layer with a rate of 0.2. The output layer was a single node that returned the unilateral hip joint moment. Hyperparameters were selected to balance training efficiency and model performance. We used the Adam optimizer with a learning rate of 0.001. We employed the mean squared error as the loss function. The batch size was set to 64. Training proceeded for up to 100 epochs, with early stopping applied if the validation loss did not improve for 10 consecutive epochs to prevent overfitting and unnecessary computations.

We conducted leave-one-subject-out cross-validation for training and testing (Fig. 1). For each iteration, the entire dataset from one subject was left out, and the remaining data was filtered based on predefined task conditions (all tasks, optimized tasks, or cyclic tasks). The filtered data was then split into 80\% training and 20\% validation sets. After training, the left-out subject's data was used for testing. Model performance was evaluated using root mean squared error (RMSE) and the coefficient of determination (R$^{2}$). RMSE assessed the accuracy of the model's predicted hip joint moment values, while R$^{2}$ evaluated the model's ability to track the characteristic pattern of the hip joint moment profile. We computed the average RMSE and R$^{2}$ for each training condition across all cross-validation folds as the final metric of the model performance.

We performed statistical analyses to evaluate the differences among the three conditions. A one-way analysis of variance (ANOVA) was conducted to test for overall differences in RMSE and R$^{2}$ values across conditions with the significance level set to 0.05. If significant differences were identified, post-hoc pairwise comparisons with a Bonferroni correction were performed to determine which specific conditions differed.

\begin{figure}[b!]
    \centering
    \includegraphics[width=0.5\textwidth]{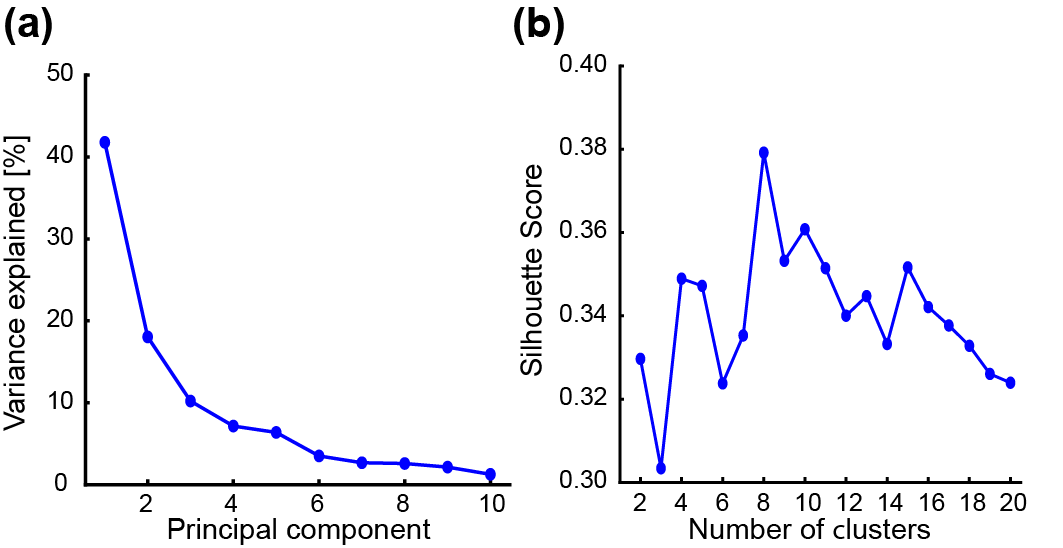}
    \caption{
        PCA and K-means Clustering Parameter Selection. (a) We used the first three principal components since they explained over 70\% of the variance. (b) The Silhouette method revealed that $K$ = 8 yielded the best silhouette score, indicating it as the optimal number of clusters for our dataset.
    }
\end{figure}

\begin{table}
\centering
\renewcommand{\arraystretch}{1.1}
\captionsetup{justification=centering, labelsep=none} 
\caption{\\Cluster Task-Weight Analysis}
\label{tab:cluster_analysis}
\resizebox{\columnwidth}{!}{
\begin{tabular}{ccccccc}
\hline
\hline
   \rule[-7pt]{0pt}{17pt}\textbf{Clusters} & \textbf{Task} & ${\frac{A}{B}}$ & ${\frac{A}{C}}$ & ${\frac{S}{11}}$ & \textbf{$w$} & \textbf{$R$} \\
\hline
\multirow{3}{*}{Cluster 1} &        Lunges &      0.172 &             0.470 &          1.000 &     0.9 & 0.07280 \\
          &  Dynamic Walk &      0.128 &             0.523 &          1.000 &     0.9 & 0.06011 \\
          &      Tire Run &      0.122 &             1.000 &          1.000 &     0.9 & 0.11000 \\
\hline
\multirow{3}{*}{Cluster 2} &   Lift Weight &      0.406 &             0.977 &          1.000 &     0.8 & 0.31715 \\
          &        Lunges &      0.165 &             0.530 &          1.000 &     0.9 & 0.07880 \\
          &     Ball Toss &      0.151 &             0.941 &          1.000 &     0.8 & 0.11365 \\
\hline
\multirow{3}{*}{Cluster 3} &     Stairs Up &      0.438 &             0.914 &          0.909 &     1.0 & 0.36387 \\
          &   Normal Walk &      0.298 &             0.493 &          1.000 &     1.0 & 0.14672 \\
          &  Incline Walk &      0.182 &             1.000 &          1.000 &     1.0 & 0.18182 \\
\hline
\multirow{3}{*}{Cluster 4} &          Jump &      0.906 &             0.755 &          1.000 &     0.9 & 0.61547 \\
          &     Curb Down &      0.047 &             0.167 &          0.273 &     0.8 & 0.00171 \\
          &  Dynamic Walk &      0.024 &             0.045 &          0.182 &     0.9 & 0.00018 \\
\hline
\multirow{3}{*}{Cluster 5} &   Normal Walk &      0.531 &             0.233 &          0.727 &     1.0 & 0.08998 \\
          &  Dynamic Walk &      0.406 &             0.295 &          0.818 &     0.9 & 0.08838 \\
          &   Stairs Down &      0.031 &             0.017 &          0.091 &     1.0 & 0.00005 \\
\hline
\multirow{3}{*}{Cluster 6} &   Stairs Down &      0.528 &             0.950 &          0.909 &     1.0 & 0.45581 \\
          & Walk Backward &      0.287 &             0.939 &          1.000 &     0.9 & 0.24268 \\
          &  Dynamic Walk &      0.056 &             0.136 &          0.364 &     0.9 & 0.00248 \\
\hline
\multirow{3}{*}{Cluster 7} &       Cutting &      0.686 &             0.795 &          1.000 &     0.8 & 0.43672 \\
          &     Curb Down &      0.137 &             0.292 &          0.455 &     0.8 & 0.01456 \\
          &       Curb Up &      0.137 &             0.292 &          0.545 &     0.8 & 0.01747 \\
\hline
\multirow{3}{*}{Cluster 8} &  Sit to Stand &      0.393 &             0.500 &          1.000 &     0.9 & 0.17679 \\
          &          Jump &      0.167 &             0.137 &          0.727 &     0.9 & 0.01497 \\
          & Turn and Step &      0.143 &             0.600 &          1.000 &     0.8 & 0.06857 \\
\hline
\hline
\end{tabular}
}
\vspace{0.5em} 
\captionsetup{justification=justified}

\caption*{\footnotesize Three tasks with the highest $R$ values in each cluster and their task-weight analysis: columns, in order, represent (1) task data proportion within the cluster, (2) task data allocated to the cluster relative to its total, (3) subject diversity rate of the task in the cluster, (4) data collection difficulty weight, and (5) representativeness score.
}

\end{table}

\begin{figure*}
    \centering
    \includegraphics[width=1.0\textwidth]{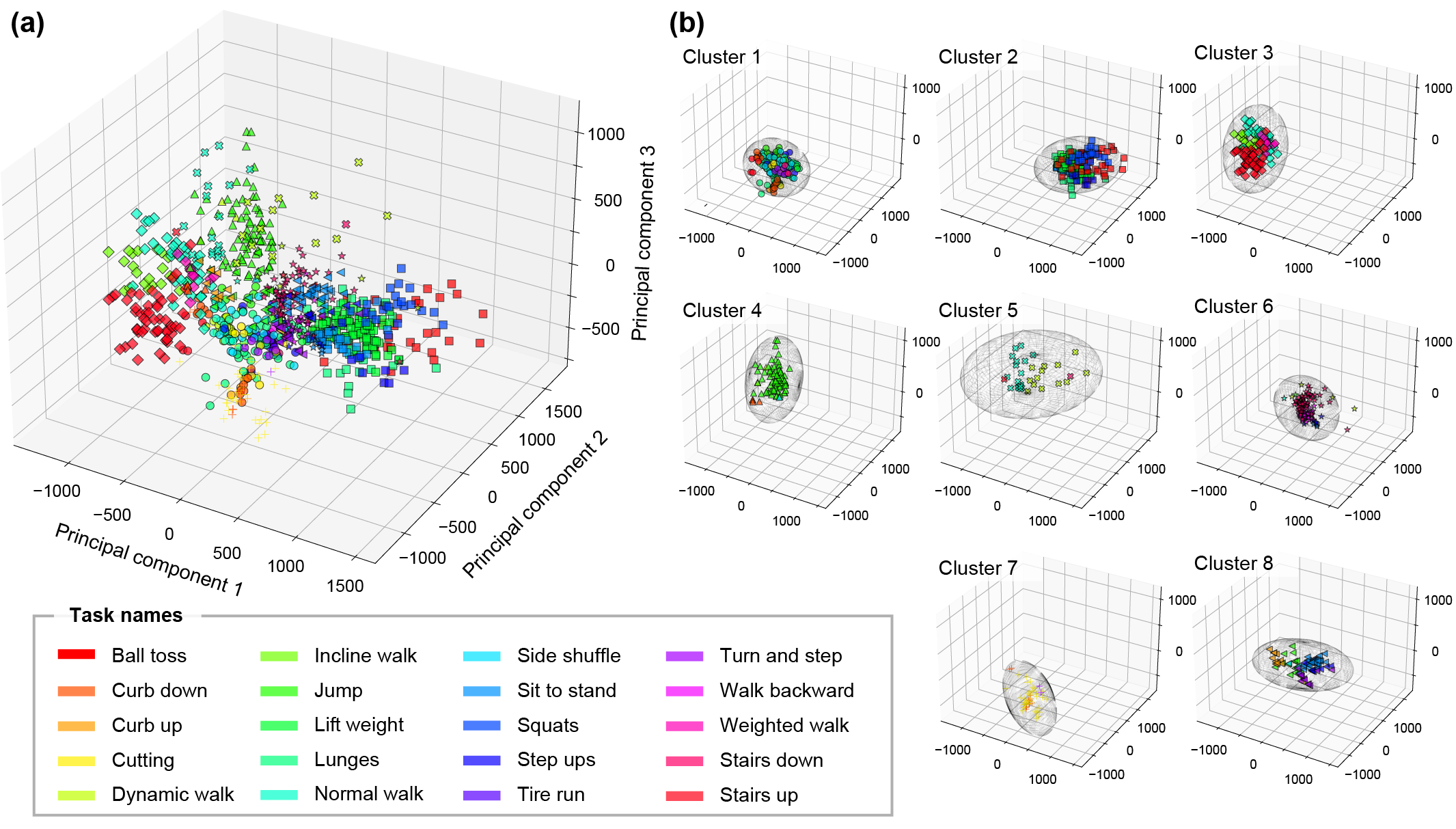}
    \caption{
         PCA and  K-means Clustering Results.
        (a) All biomechanical feature data represented in the reduced dimensions. Each task mode is colored differently and each cluster is marked in a different marker shape. (b) Separate plots for each cluster, with the ellipsoid representing the space of 99\% confidence for each cluster.
        }
\end{figure*}

\section{Results}
\subsection{PCA and Clustering Results}
The first three principal components accounted for 41.79\%, 18.04\%, and 10.21\% of the variance, respectively (Fig. 2a). By projecting the original data onto these top three principal components, a three-dimensional representation of the biomechanical latent space of a total of 70.04\% of explained variance was generated (Fig. 3a). This reduced-dimensional space served as the foundational coordinate system for our clustering analysis.

The optimal cluster number for our dataset was eight with a silhouette score of 0.38 (Fig. 2b). Using the clustering result (Fig. 3b) and task-weight analysis, the representativeness of each task, $R$, was computed (Table 1). The task with the highest $R$ score in each cluster was selected as the representative task for the cluster. Among these eight tasks, three were cyclic tasks (normal Walk, stairs up, and stairs down) and five were non-cyclic tasks (lunges, jump, cutting, sit to stand, and lift weight). These tasks were our training dataset for the optimized model. The other two conditions were trained with 1) all tasks, which included all 20 different tasks and 2) cyclic tasks, which included all eight cyclic tasks.

\begin{figure}
    \centering
    \includegraphics[width=0.5\textwidth]{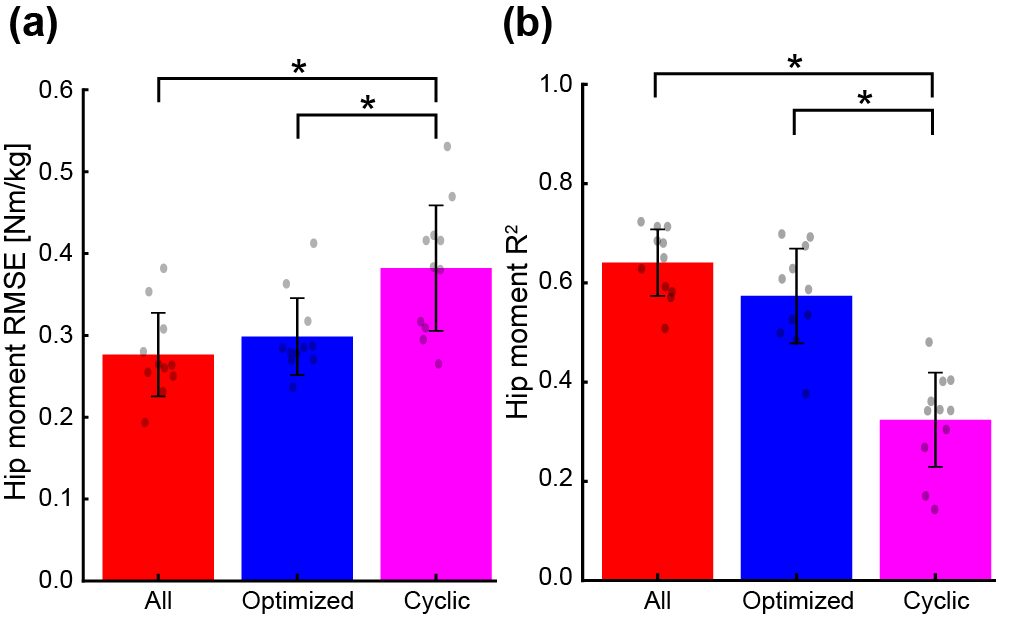}
    \caption{
       Model Performance. (a) The all-task and optimized-task models outperformed the cyclic-task model, with significantly lower root mean squared errors (RMSEs). (b) Similarly, the optimized-task model showed higher R$^{2}$ compared to the cyclic task model, but similar performance compared to the all-task model. Gray dots represent values from 11 models per condition. * indicates a statistical difference from the cyclic tasks model (p<0.05). 
        }
\end{figure}

\begin{figure}
    \centering
    \includegraphics[width=0.50\textwidth]{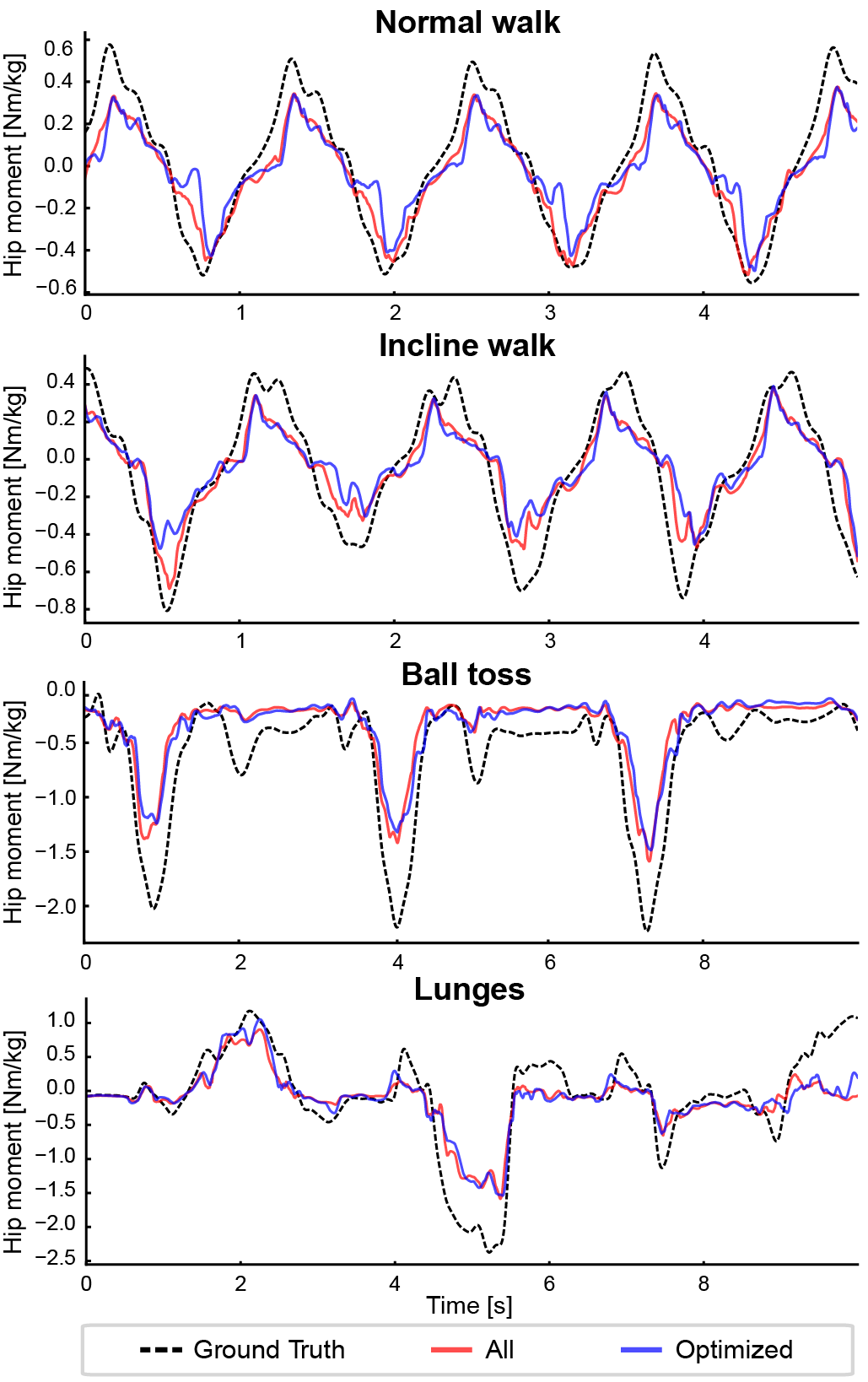}
    \caption{
        Qualitative Performance Evaluation. Ground truth and estimated hip joint moments for the optimized-task (blue) and all-task (red) models on two cyclic and two non-cyclic tasks using leave-out subject data with median RMSE performance. Ground truth is shown as a black dashed line. Both models follow the ground truth trend, but peak moment estimates show a ceiling effect.
        }
\end{figure}

\subsection{Neural Network Performance}
The optimized-task model performed better than the cyclic-task model and as well as the all-task model. The optimized model achieved an average RMSE of 0.30$\pm$0.05 Nm/kg and R$^{2}$ of 0.57$\pm$0.10 (Fig. 4). The all task model had an average RMSE of 0.28$\pm$0.05 Nm/kg and R$^{2}$ of 0.64$\pm$0.07. The cyclic task model presented an average RMSE of 0.38$\pm$0.08 Nm/kg and R$^{2}$ of 0.32$\pm$0.09. The models trained with all tasks and optimized tasks had lower RMSEs compared to the model trained with all cyclic tasks, by 27.11$\pm$6.18\% and 20.60$\pm$9.80\%, respectively (p<0.05). The optimized-task model was not different, by both RMSE and R$^{2}$, compared to the all-task model. The ground truth versus predicted hip joint moments of the optimized task model and the all task model are depicted for two cyclic (normal walk and incline walk) and two non-cyclic (ball toss and lunges) tasks, serving as representative examples of each task type (Fig. 5).

\section{Discussion}
As hypothesized, the model trained on the optimized task set achieved comparable performance to the model trained on all tasks (i.e., the entire dataset) in hip joint moment estimation (Fig. 4). This finding validates the effectiveness of the task set optimization approach we used. Notably, the optimized task set outperformed the cyclic task data in RMSE and R$^{2}$ metrics, even though both sets contained the same number of tasks. This result highlighted the importance of including non-cyclic tasks in training data. It also presented the advantage of task optimization over relying solely on cyclic tasks for training a biological joint moment estimator.

The comparison of model predictions supports our argument that the optimized task model performs comparably to the all task model (Fig. 5). Both models were tested on a leave-out subject whose data showed median RMSE performance with the optimized task model. Despite its median performance, the model produced predictions nearly equivalent to those of the all task model. Furthermore, when predictions were poorly executed, both models underperformed. This suggests that these difficulties are not stemming from the training tasks but from the limitation of using an FCNN and the inherent challenges of predicting non-cyclic tasks using only kinematic data. Overall, these results imply that developing a user-independent hip joint moment estimator is feasible with reduced task data. This approach could streamline future exoskeleton controller design by reducing data collection and training demands, facilitating the deployment of hip exoskeletons for daily activities.

Compared to a previous hip joint torque estimation using the same dataset, our optimized task model demonstrated comparable accuracy \cite{Scherpereel_SciData_2023}. The study trained temporal convolutional networks (TCN) with kinematic sensor inputs (pelvis and thigh IMUs, hip angle, and velocity) and achieved an RMSE of 0.280 Nm/kg and an R$^{2}$ of 0.56 \cite{Scherpereel_SciData_2023}. Using an FCNN, our model achieved similar performance, likely due to the exclusion of seven non-cyclic tasks with atypical movements. Although these excluded tasks were more challenging to predict accurately, the task set we included remains diverse compared to datasets limited to only treadmill or cyclic tasks. This comparison indicated the effectiveness of our task optimization and demonstrated that model performance is maintained even with fewer task data.

Furthermore, our approach of grouping tasks into clusters based on biomechanical features proved successful, offering a foundation for interpretable task relationships.
The previous study applied a forward task selection algorithm and suggested that their ten tasks were sufficient to achieve optimal model performance \cite{Scherpereel_SciData_2023}. Three of their tasks (sit-to-stand, normal walk, and jumping) overlapped with our optimized tasks. Their inclusion of some tasks that we excluded may have caused these distinct results. Nevertheless, this comparison highlighted the potential of our method to optimize task selection for model training through a more interpretable and biomechanics-grounded approach. From our clustering result, we can see that tasks were grouped with similarly performed tasks (Table 1). For instance, Cluster 3 predominantly included cyclic tasks focused on ascending movements (stairs up, incline walk, and normal walk), while Cluster 2 comprised non-cyclic and propulsive tasks (lift weight, lunges, and ball toss). These results demonstrated that our method provided a more intuitive understanding of task relationships compared to black box, performance-driven approaches. Thus, our proposed approach was not only cost-efficient but also more interpretable, making it a practical alternative for task selection in model training.

While our study presented promising results, it has several limitations. 
First, compared to the previous studies that used more complex models like long short-term memory (LSTM) networks \cite{LSTM} or TCN \cite{Molinaro_BioRob_2020}, we opted for a simple FCNN. While the neural network architecture was intentionally kept simple to serve as a controlled variable in this study, a more complex model might better capture the intricate relationships between tasks within each cluster. Such models could also more effectively highlight the impact of task selection optimization. Future research could explore architectures like LSTM or TCN and fine-tune such models to investigate their potential advantages in this context. 
Second, the dataset used in this study exhibited an imbalanced number of trials for each task, which could influence the outcomes of the dimensionality reduction and clustering processes. Although a task-weight analysis formula was employed to mitigate these effects, ensuring an equal number of data points for each task in future datasets could further enhance the quality and reliability of the PCA and k-means clustering. Finally, this study used data only from healthy participants. Our approach relied on biomechanical profile similarities, which limits its effectiveness in individuals with significant differences, such as patient populations. Severe differences (e.g., crouch gait or stiff knee) may degrade the task optimization process and, consequently, model performance. Further research with more diverse subjects is needed to refine the approach for patient applications.

\section{Conclusion}
In this work, we presented a method to optimize task selection for a user-independent hip joint moment estimator. We identified the most representative tasks for model training by leveraging dimensionality reduction and clustering techniques on biomechanical features. The model trained on these optimized tasks achieved comparable performance to the model trained on the full dataset. These findings underscore the robustness of our approach and provide a pathway for future research on hip exoskeleton controller development with minimal training efforts.

\bibliographystyle{ieeetr}
\bibliography{References}

\end{document}